\documentclass[11pt]{article}%
\usepackage{amsmath}
\usepackage{amsfonts}
\usepackage{amssymb}
\usepackage{graphicx}
\usepackage{algorithmic}
\usepackage{algorithm}%
\providecommand{\U}[1]{\protect\rule{.1in}{.1in}}

\begin{document}

\title{Improved Anomaly Detection by Using the Attention-Based Isolation Forest}
\author{Lev V. Utkin, Andrey Y. Ageev and Andrei V. Konstantinov\\Peter the Great St.Petersburg Polytechnic University\\St.Petersburg, Russia\\e-mail: lev.utkin@gmail.com, andreyageev1@mail.ru, andrue.konst@gmail.com}
\date{}
\maketitle

\begin{abstract}
A new modification of Isolation Forest called Attention-Based Isolation Forest
(ABIForest) for solving the anomaly detection problem is proposed. It
incorporates the attention mechanism in the form of the Nadaraya-Watson
regression into the Isolation Forest for improving solution of the anomaly
detection problem. The main idea underlying the modification is to assign
attention weights to each path of trees with learnable parameters depending on
instances and trees themselves. The Huber's contamination model is proposed to
be used for defining the attention weights and their parameters. As a result,
the attention weights are linearly depend on the learnable attention
parameters which are trained by solving the standard linear or quadratic
optimization problem. ABIForest can be viewed as the first modification of
Isolation Forest, which incorporates the attention mechanism in a simple way
without applying gradient-based algorithms. Numerical experiments with
synthetic and real datasets illustrate outperforming results of ABIForest. The
code of proposed algorithms is available.

Keywords: anomaly detection, attention mechanism, Isolation Forest,
Nadaraya-Watson regression, quadratic programming, contamination model

\end{abstract}

\section{Introduction}

One of the important machine learning problems is the novelty or anomaly
detection problem which aims to detect abnormal or anomalous instances. This
problem can be regarded as a challenging task because there is no a strong
definition of anomalous instance and the anomaly itself depends on a certain
application. Another difficulty which defines the challenge of the problem is
that anomalies usually seldom appear and this fact leads to highly imbalanced
training sets. Moreover, it is difficult to define a boundary between the
normal and anomalous observations \cite{Chalapathy-Chawla-22}. Due to
importance of the anomaly detection problem in many applications, a huge
amount of papers covering anomaly detection tasks and studying various aspects
of the anomaly detection have been published in the last decades. Many
approaches to solving the anomaly detection problem are analyzed in
comprehensive survey papers
\cite{Chalapathy-Chawla-22,Boukerche-etal-20,Braei-Wagner-20,Chandola-etal-09,Farizi-etal-21,Fauss-etal-21,Guansong-Pang-etal-22,Jingkang-Yang-etal-22,Pang-Cao-Aggarwal-21,Ruff-etal-21,Wang-Bah-Hammad-19}%
.

According to \cite{Chalapathy-Chawla-22,Aggarwal-13}, anomalies also referred
to as abnormalities, deviants, or outliers can be viewed as data points which
are located further away from the bulk of data points that are referred to as
normal data.

Various approaches to solving the anomaly detection problem can be divided
into several groups \cite{Ruff-etal-21}. The first group consists of the
probabilistic and density estimation models. It includes the classic density
estimation models, energy-based models, neural generative models
\cite{Ruff-etal-21}. The second large group deals with the one-class
classification models. This group includes the well-known one-class
classification SVMs \cite{Campbell-Bennett-2001,Scholkopf-2001,Tax-Duin-2004}.
The third group includes reconstruction-based models which detect anomalies by
reconstructing the data instances. The well-known models from this group are
autoencoders which incorrectly reconstruct anomalous instances such that the
distance between the instance and its reconstruction is larger than a
predefined threshold which is usually regarded as a hyperparameter of the model.

The next group contains distance-based anomaly detection models. One of the
most popular and effective models from the group is the Isolation Forest
(iForest) \cite{Liu-Kai-Zhou-08,Liu-Ting-Zhou-2012} which is a model for
detecting anomalous points relative to a certain data distribution. According
to iForest, anomalies are detected using isolation which measures how far an
instance is from the rest of instances. iForest can be regarded as a tool
implementing the isolation. It has the linear time complexity and works well
with large amounts of data. The core idea behind iForest is the tendency for
anomalous instances in a dataset to be more easily separated from the rest of
the sample (isolated) compared to normal instances. To isolate a data point,
the algorithm recursively creates sample partitions by randomly choosing an
attribute and then randomly choosing a split value for the attribute between
the minimum and maximum values allowed for that attribute. The recursive
partition can be represented by a tree structure called an isolation tree,
while the number of partitions needed to isolate a point can be interpreted as
the length of the path within the tree to the end node, starting from the
root. Anomalous instances are those with a shorter path length in the tree
\cite{Liu-Kai-Zhou-08,Liu-Ting-Zhou-2012}.

In order to improve iForest, we propose to modify it by using the attention
mechanism which can automatically distinguish the relative importance of
instances and weigh them for improving the overall accuracy of iForest. The
attention mechanism has been successfully applied to many applications,
including the natural language processing models, the computer vision area,
etc. Comprehensive surveys of properties and forms of the attention mechanism
and transformers can be found in
\cite{Chaudhari-etal-2019,Correia-Colombini-21a,Correia-Colombini-21,Lin-Wang-etal-21,Niu-Zhong-Yu-21}%
.

The idea to apply the attention mechanism to iForest stems from the
attention-based random forest (ABRF) models proposed in
\cite{Utkin-Konstantinov-22} where attention is implemented in the form of the
Nadaraya-Watson regression \cite{Nadaraya-1964,Watson-1964} by assigning
attention weights to leaves of trees in a specific way such that the weights
depend on trees and instances. The attention learnable parameters in ABRF are
trained by solving the standard quadratic optimization problem with linear
constraints. It turns out that this idea to consider the random forest as the
Nadaraya-Watson regression \cite{Nadaraya-1964,Watson-1964} can be extended to
iForest taking into account the iForest peculiarities which differ it from the
random forest. According to the original iForest, the isolation measure is
estimated as the mean value of the path lengths over all trees in the forest.
However, we can replace the averaging of the path lengths with the
Nadaraya-Watson regression where the path length of an instance in each tree
can be regarded as a prediction in the regression (the \textit{value} in terms
of the attention mechanism \cite{Bahdanau-etal-14}), and weights (the
attention weights) depend on the corresponding tree and the instance (the
\textit{query} in terms of the attention mechanism \cite{Bahdanau-etal-14}).
In other words, the final prediction of the expected path length in accordance
with the Nadaraya-Watson regression is a weighted sum of path lengths over all
trees. Weights of path lengths have learnable parameters (the learnable
attention parameters) which can be computed by minimizing a loss function of a
specific form. We aim to reduce the optimization problem to the quadratic
programming problem or linear programming problem which has many algorithms
for solving. In order to achieve this aim, the Huber's $\epsilon
$-contamination model \cite{Huber81} is proposed to be used for computing the
learnable attention parameters. The contamination model allows us to represent
attention weights in the form of a linear combination of the softmax operation
and learnable parameters with contamination parameter $\epsilon$, which can be
viewed as probabilities. As a result, the loss function for computing
learnable parameters is linear with linear constraints on the parameters as
probabilities. After adding the $L_{2}$ regularization term, the optimization
problem for computing attention weights becomes to be quadratic one.

Our contributions can be summarized as follows:

\begin{enumerate}
\item A new modification of iForest called Attention-Based Isolation Forest
(ABIForest) incorporating the attention mechanism in the form of the
Nadaraya-Watson regression for improving solution of the anomaly detection
problem is proposed.

\item The algorithm of computing attention weights is reduced to solving the
linear or quadratic programming problems due to applying the Huber's
$\epsilon$-contamination model. Moreover, we propose to use the hinge-loss
function to simplify the optimization problem. Contamination parameter
$\epsilon$ is regarded as a tuning hyperparameter.

\item Numerical experiments with synthetic and real datasets are performed for
studying ABIForest. They demonstrate outperforming results for most datasets.
The code of proposed algorithms can be found in https://github.com/AndreyAgeev/Attention-based-isolation-forest.
\end{enumerate}

The paper is organized as follows. Related work can be found in Section 2.
Brief introductions to the attention mechanism, the Nadaraya-Watson regression
and iForest are given in Section 3. The proposed ABIForest model is considered
in Section 4. Numerical experiments with synthetic and real datasets
illustrating peculiarities of ABIForest and its comparison with iForest are
provided in Section 5. Concluding remarks discussing advantages and
disadvantages of ABIForest can be found in Section 6.

\section{Related work}

\textbf{Attention mechanism}. The attention mechanism can be viewed as an
effective method for improving the performance of a large variety of machine
learning models. Therefore, there are many different types of attention
mechanisms depending on their applications and models where attention
mechanisms are incorporated. The term \textquotedblleft
attention\textquotedblright\ was introduced by Bahdanau et al.
\cite{Bahdanau-etal-14}. Following this paper, a huge amount of models based
on the attention mechanism can be found in the literature. There are also
several types of attention mechanisms \cite{Sawant-Singh-20}, including soft
and hard attention mechanisms \cite{Xu-Ba-etal-15}, the local and global
attention \cite{Luong-Pham-Manning-15}, self-attention \cite{Vaswani-etal-17},
multi-head attention \cite{Vaswani-etal-17}, hierarchical attention
\cite{Yang-Yang-etal-16}. It is difficult to consider all papers devoted to
the attention mechanisms and its applications. Comprehensive surveys
\cite{Chaudhari-etal-2019,Correia-Colombini-21a,Correia-Colombini-21,Lin-Wang-etal-21,Niu-Zhong-Yu-21,Liu-Huang-etal-21}
cover a large part of available models and modifications of the attention mechanisms.

Most attention models are implemented as parts of neural networks. In order to
extend a set of attention models, several random forest models incorporated
with the attention mechanism were proposed in
\cite{Utkin-Konstantinov-22,Utkin-Konstantinov-22c,Utkin-Konstantinov-22d}.
The gradient boosting machine added by the attention mechanism was presented
in \cite{Konstantinov-Utkin-22d}.

\textbf{Anomaly detection with attention}. A wide set of machine learning
tasks include anomaly detection problems. Therefore, many methods and models
have been developed to address them
\cite{Chalapathy-Chawla-22,Boukerche-etal-20,Braei-Wagner-20,Chandola-etal-09,Farizi-etal-21,Fauss-etal-21,Guansong-Pang-etal-22,Jingkang-Yang-etal-22,Pang-Cao-Aggarwal-21,Ruff-etal-21,Wang-Bah-Hammad-19}%
. One of the tools for solving the anomaly detection problems is the attention
mechanism. Monotonic attention based autoencoders was proposed in
\cite{Kundu-etal-20} as an unsupervised learning technique to detect the false
data injection attacks. Anomaly detection method based on the Siamese network
with an attention mechanism for dealing with small datasets was proposed in
\cite{Takimoto-etal-22}. The so-called residual attention network that employs
the attention mechanism and residual learning to improve classification
efficiency and accuracy was presented in \cite{Lei-Xia-etal-23}. The graph
anomaly detection algorithm based on the attention-based deep learning to
assist the audit process was provided in \cite{Yu-Zha-etal-22}. Madan et al.
\cite{Madan-etal-22} presented a novel self-supervised masked convolutional
transformer block that comprises the reconstruction-based functionality.
Integration of the reconstruction-based functionality into a novel
self-supervised predictive architectural building block was considered in
\cite{Ristea-etal-22}. Huang et al. \cite{Huang-Xu-Wang-etal-22} improved the
efficiency and effectiveness of anomaly detection and localization at
inference by using a progressive mask refinement approach that progressively
uncovers the normal regions and finally locates the anomalous regions. A novel
self-supervised framework for multivariate time-series anomaly detection via a
graph attention network was proposed in \cite{Hang_Zhao-etal-20}. It can be
seen from the above works that the idea to apply attention in models solving
the anomaly detection problem was successfully implemented. However, the
attention was used in the form of components of neural networks. There are no
forest-based anomaly detection models which use the attention mechanism.

\textbf{iForest.} iForest \cite{Liu-Kai-Zhou-08,Liu-Ting-Zhou-2012} can be
viewed as one of the important and effective methods for solving novelty and
anomaly detection problems. Therefore, many modifications of the method have
been developed \cite{Farizi-etal-21} to improve it. A weighted iForest and
Siamese Gated Recurrent Unit algorithm architecture which provides a more
accurate and efficient method for outlier detection of data is considered in
\cite{Junfeng_Wang-etal-22}. Hariri et al. \cite{Hariri-etal-21} proposed an
extension of the iForest, named Extended Isolation Forest, which resolves
issues with assignment of anomaly score to given data points. A theoretical
framework that describes the effectiveness of isolation-based approaches from
a distributional viewpoint was studied in \cite{Buschjager-etal-20}. Lesouple
et al. \cite{Lesouple-etal-21} presented a generalized isolation forest
algorithm which generates trees without any empty branch, which significantly
improves the execution times. The k-Means-Based iForest was developed by
Karczmarek et al. \cite{Karczmarek-etal-20}. This modification of iForest
allows to build a search tree based on many branches in contrast to the only
two considered in the original method. Another modification, called the Fuzzy
Set-Based Isolation Forest was proposed in \cite{Karczmarek-etal-20a}. A
probabilistic generalization of iForest was proposed in
\cite{Tokovarov-Karczmarek-22}, which is based on nonlinear dependence of a
segment-cumulated probability from the length of segment. A robust anomaly
detection method called the similarity-measured isolation forest was developed
by Li et al. \cite{Li-Guo-Gao-Li-21} to detect abnormal segments in monitoring
data. A novel hyperspectral anomaly detection method with kernel Isolation
Forest was proposed in \cite{Li-Zhang-Duan-Kang-20}. The method is based on an
assumption that anomalies rather than background can be more susceptible to
isolation in the kernel space. An improved computational framework which
allows us to seek the most separable attributes and spot corresponding
optimized split points effectively was presented in \cite{Liu-Liu-Ma-Gao-18}.
Staerman et al. \cite{Staerman-etal-19} introduced the so-called Functional
Isolation Forest which generalizes iForest to the infinite dimensional
context, i.e., the model deals with functional random variables that take
their values in a space of functions. Xu et al. \cite{Xu-Peng-etal-22}
proposed the Deep Isolation Forest which is based on an isolation method with
arbitrary (linear/non-linear) partition of data implemented by using neural networks.

The above works is only a part of many extensions and modifications of iForest
developed due to excellent properties of the method. However, to the best of
our knowledge, there are no works considering approaches to incorporating the
attention mechanism into iForest.

\section{Preliminaries}

\subsection{Attention mechanism as the Nadaraya-Watson regression}

If to consider the attention mechanism as a method for enhancing accuracy of
iForest for the anomaly detection problem solution, then it allows us to
automatically distinguish the relative importance of features, instances and
isolation trees. According to \cite{Chaudhari-etal-2019,Zhang2021dive}, the
original idea of attention can be understood from the statistical point of
view applying the Nadaraya-Watson kernel regression model
\cite{Nadaraya-1964,Watson-1964}.

Given $n$ instances $\mathcal{D}=\{(\mathbf{x}_{1},y_{1}),...,(\mathbf{x}%
_{n},y_{n})\}$, in which $\mathbf{x}_{i}=(x_{i1},...,x_{id})\in\mathbb{R}^{d}$
is a feature vector involving $m$ features and $y_{i}\in\mathbb{R}$ represents
the regression outputs, the task of regression is to construct a regressor
$f:\mathbb{R}^{m}\rightarrow\mathbb{R}$ which can predict the output value
$\tilde{y}$ of a new observation $\mathbf{x}$, using available data $S$. The
similar task can be formulated for the classification problem.

The original idea behind the attention mechanism is to replace the simple
average of outputs $\tilde{y}=n^{-1}\sum_{i=1}^{n}y_{i}$ for estimating the
regression output $y$, corresponding to a new input feature vector
$\mathbf{x}$ with the weighted average, in the form of the Nadaraya-Watson
regression model \cite{Nadaraya-1964,Watson-1964}:%
\begin{equation}
\tilde{y}=\sum_{i=1}^{n}\alpha(\mathbf{x},\mathbf{x}_{i})y_{i},
\end{equation}
where weight $\alpha(\mathbf{x},\mathbf{x}_{i})$ conforms with relevance of
the $i$-th instant to the vector $\mathbf{x}$, i.e., it is defined in
agreement with the corresponding input $\mathbf{x}_{i}$ locations relative to
the input variable $\mathbf{x}$ (the closer an input $\mathbf{x}_{i}$ to the
given variable $\mathbf{x}$, the greater $\alpha(\mathbf{x},\mathbf{x}_{i})$).

In terms of the attention mechanism \cite{Bahdanau-etal-14}, vectors
$\mathbf{x}$, $\mathbf{x}_{i}$ and outputs $y_{i}$ are called as the
\textit{query}, \textit{keys} and \textit{values}, respectively. Weight
$\alpha(\mathbf{x},\mathbf{x}_{i})$ is called as the attention weight.

The attention weights $\alpha(\mathbf{x},\mathbf{x}_{i})$ can be defined by a
normalized kernel $K$ as:
\begin{equation}
\alpha(\mathbf{x},\mathbf{x}_{i})=\frac{K(\mathbf{x},\mathbf{x}_{i})}%
{\sum_{j=1}^{n}K(\mathbf{x},\mathbf{x}_{j})}.
\end{equation}

For the Gaussian kernel with parameter $\omega$, the attention weights are
represented through the softmax operation as:
\begin{equation}
\alpha(\mathbf{x},\mathbf{x}_{i})=\sigma\left(  -\frac{\left\Vert
\mathbf{x}-\mathbf{x}_{i}\right\Vert ^{2}}{\omega}\right)  .
\end{equation}

In order to enhance the attention capability, weights are added by trainable
parameters. Several definitions of attention weights and attention mechanisms
have been proposed. The most popular definitions are the additive attention
\cite{Bahdanau-etal-14}, the multiplicative or dot-product attention
\cite{Luong-etal-2015,Vaswani-etal-17}.

\subsection{Isolation forest}

In this subsection, the main definitions of iForest are provided in accordance
with results given in \cite{Liu-Kai-Zhou-08,Liu-Ting-Zhou-2012}. Suppose that
there is a dataset $\mathcal{D}=\{\mathbf{x}{_{1},...,\mathbf{x}_{n}}\}$
consisting of $n$ instances, where $\mathbf{x}_{i}=(x_{i1},...,x_{id}%
)\in\mathbb{R}^{d}$ is a feature vector. The isolation tree is built by using
a randomly generated subset $\mathcal{D}^{\ast}$ of the dataset $\mathcal{D}$.
The dataset $\mathcal{D}^{\ast}$ splits into two subsets to define a random
node as follows. A feature is randomly selected by generating the random value
$q$ from the set $\{1,...,d\}$. Then a split value $p$ is randomly selected
from interval $[\min_{i=1,...,n}x_{iq},\max_{i=1,...,n}x_{iq}]$. Having $p$
and $q$, the subset $\mathcal{D}^{\ast}$ is recursively divided at each node
by using the feature number $q$ and the split value $p$ into two parts: the
left branch corresponding to the set with $x_{iq}\leq p$ and the right branch
corresponding to the set with $x_{iq}>p$. Thus generated values $q$ and $p$
determine whether the data points at a node are sent down the left or the
right branch. The above conditions determine the subsequent child nodes for a
split node. The division stops in accordance with a rule, for example, when a
branch contains a single point or when some limited depth of the tree is
reached. The process of the isolation tree building begins again with a new
random subsample to build another randomized tree. After building a forest
consisting of $T$ trees, the training process is complete.

In the $k$-th isolation tree, an instance $\mathbf{x}$ is isolated on one of
the outer nodes such that a path of length $h_{k}(\mathbf{x})$ can be
associated with this instance, which is defined as a number of nodes that
$\mathbf{x}$ goes from the root node to the leaf. Anomalous instances are
those with a shorter path length in the tree. This conclusion is motivated by
the fact that normal instances are more concentrated than anomalies and thus
require more nodes to be isolated. By having the trained $T$ trees, i.e., the
isolated forest, we can estimate the isolation measure as the expected path
length $E[h(\mathbf{x})]$ which is computed as the mean value of the path
lengths over all trees in the forest. By having the expected path length
$E[h(\mathbf{x})]$, an anomaly score is defined as
\begin{equation}
s(\mathbf{x},n)=2^{-\frac{E(h(\mathbf{x}))}{c(n)}}, \label{IForest_10}%
\end{equation}
where $c(n)$ is is the normalizing factor defined as the average value of
$h(\mathbf{x})$ for a dataset of size $n$, which is computed as%

\begin{equation}
c(n)=2H(n-1)-\frac{2(n-1)}{n}.
\end{equation}

Here $H(n)$ is the $n$-th harmonic number estimated from ${H(n)=\ln
(n)+}{\delta}$, where $\delta\approx0.577216$ is the Euler-Mascheroni
constant. If $n=2$, then $c(n)=1$.

The higher the value of $s(\mathbf{x},n)$ (closer to $1$), the more likely the
instance $\mathbf{x}$ is anomalous. If we introduce a threshold $\tau
\in\lbrack0.1]$, then condition $s(\mathbf{x},n)>$ $\tau$ indicates that
instance $\mathbf{x}$ is detected as an anomaly. If condition $s(\mathbf{x}%
,n)\leq\tau$ is valid, then instance $\mathbf{x}$ is likely normal. The
threshold $\tau$ in the original iForest is taken $0.5$.

\section{Attention-Based Isolation Forest}

It should be noted that the expected path length $E[h(\mathbf{x})]$ in the
original iForest is computed as the mean value of the path lengths
$h_{k}(\mathbf{x})$ of trees:%
\begin{equation}
E[h(\mathbf{x})]=\frac{1}{T}\sum_{k=1}^{T}h_{k}(\mathbf{x}).
\label{IForest_20}%
\end{equation}

This method for computing the expected path length does not take into account
the possible relationship between an instance and each isolation tree, the
possible difference between trees. Ideas behind the attention-based RF
\cite{Utkin-Konstantinov-22} can also be applied to iForest. Therefore, our
next task is to incorporate the attention mechanism into iForest.

\subsection{Keys-values and the query in iForests}

First, we can point out that the outcome of each isolation tree is the path
length $h_{k}(\mathbf{x})$, $k=1,...,n$. This implies that this outcome can be
regarded as the value in the attention mechanism. Second, we define the query
and keys in iForest. Suppose that the feature vector $\mathbf{x}$ falls into
the $i$-th leaf of the $k$-th tree. Let $\mathcal{J}_{i}^{(k)}$ be a set of
indices of $n_{i}^{(k)}$ training instances $\mathbf{x}_{j}$ which also felt
into the same leaf. A distance between vector $\mathbf{x}$ and all vectors
$\mathbf{x}_{j}$, $j\in\mathcal{J}_{i}^{(k)}$, shows how vector $\mathbf{x}$
is in agreement with corresponding vectors $\mathbf{x}_{j}$, how it is close
to vectors $\mathbf{x}_{j}$ from the same leaf. If the distance is small, then
we can conclude that vector $\mathbf{x}$ is well performed by the $k$-th tree.
The distance between vector $\mathbf{x}$ and all vectors $\mathbf{x}_{j}$,
$j\in\mathcal{J}_{i}^{(k)}$, can be represented as a distance between vector
$\mathbf{x}$ and the mean values of all vectors $\mathbf{x}_{j}$ with indices
$j\in\mathcal{J}_{i}^{(k)}$. The mean vector of $\mathbf{x}_{j}$ with indices
$j\in\mathcal{J}_{i}^{(k)}$ can be viewed as a characteristic of the
corresponding path, i.e., this vector characterizes a group of instances which
fall into the corresponding leaf. Hence, the mean vector shows how vector
$\mathbf{x}$ is in agreement with this group. If we denote the mean value of
$\mathbf{x}_{j}$, $j\in\mathcal{J}_{i}^{(k)}$ as $\mathbf{A}_{k}(\mathbf{x)}$,
then there holds
\begin{equation}
\mathbf{A}_{k}(\mathbf{x)}=\frac{1}{n_{i}^{(k)}}\sum_{j\in\mathcal{J}%
_{i}^{(k)}}\mathbf{x}_{j}. \label{RF_Att_20}%
\end{equation}

We omit the index $j$ in $\mathbf{A}_{k}(\mathbf{x)}$ because the instance
$\mathbf{x}$ can fall only into one leaf of each tree.

Vectors $\mathbf{A}_{k}(\mathbf{x)}$ and $\mathbf{x}$ can be regarded as the
key and the query, respectively. Then (\ref{IForest_20}) can be rewritten by
using the attention weights $\alpha\left(  \mathbf{x},\mathbf{A}%
_{k}(\mathbf{x)},\mathbf{w}\right)  $ as follows:
\begin{equation}
E[h(\mathbf{x})]=\sum_{k=1}^{T}\alpha\left(  \mathbf{x},\mathbf{A}%
_{k}(\mathbf{x)},\mathbf{w}\right)  \cdot h_{k}(\mathbf{x}),
\label{IForest_25}%
\end{equation}
where $\alpha\left(  \mathbf{x},\mathbf{A}_{k}(\mathbf{x)},\mathbf{w}\right)
$ conforms with relevance of \textquotedblleft mean instance\textquotedblright%
\ $\mathbf{A}_{k}(\mathbf{x)}$ to vector $\mathbf{x}$ and satisfies condition
\begin{equation}
\sum_{k=1}^{T}\alpha\left(  \mathbf{x},\mathbf{A}_{k}(\mathbf{x)}%
,\mathbf{w}\right)  =1,\ \alpha\left(  \mathbf{x},\mathbf{A}_{k}%
(\mathbf{x)},\mathbf{w}\right)  \geq0,\ k=1,...,T.
\end{equation}

We have replaced the expected path length (\ref{IForest_20}) with the weighted
sum of path lengths (\ref{IForest_25}) such that weights $\alpha$ depend on
$\mathbf{x}$, mean vector $\mathbf{A}_{k}(\mathbf{x)}$ and vector of
parameters $\mathbf{w}$. Vector $\mathbf{w}$ in attention weights represents
trainable attention parameters. Their values depend on the dataset and on the
isolation tree properties. If we return to the Nadaraya-Watson kernel
regression model, then the expected path length $E[h(\mathbf{x})]$ can be
viewed as the regression output, path lengths $h_{k}(\mathbf{x})$ of all trees
for query $\mathbf{x}$ are predictions (values in terms of the attention
mechanism \cite{Bahdanau-etal-14}).

Suppose that the trainable parameters $\mathbf{w}$ belong to a set
$\mathcal{W}$. Then they can be found by solving the following optimization
problem:
\begin{equation}
\mathbf{w}_{opt}=\arg\min_{\mathbf{w\in}\mathcal{W}}~\sum_{s=1}^{n}L\left(
E[h(\mathbf{x}_{s})],\mathbf{x}_{s},\mathbf{w}\right)  . \label{RF_Att_49}%
\end{equation}

Here $L\left(  E[h(\mathbf{x}_{s})],\mathbf{x}_{s},\mathbf{w}\right)  $ is the
loss function whose definition as well as the definition of $\alpha\left(
\mathbf{x},\mathbf{A}_{k}(\mathbf{x)},\mathbf{w}\right)  $ are the next tasks.

\subsection{Loss function and attention weights}

First, we reformulate the decision rule ($s(\mathbf{x},n)>$ $\tau$) for
determining anomalous instances by establishing a similar condition for
$E[h(\mathbf{x})]$. Suppose that $\gamma$ is a threshold such that condition
$E[h(\mathbf{x})]\leq\gamma$ indicates that instance $\mathbf{x}$ is detected
as an anomaly. Then it follows from (\ref{IForest_10}) that $\gamma$ can be
expressed through threshold $\tau$ as:
\begin{equation}
\gamma=-c(n)\cdot\log_{2}(\tau).
\end{equation}

Hence, we can write the decision rule about the anomaly as follows:
\begin{equation}
\text{decision}=%
\begin{cases}
\text{anomalous}, & \text{if }E[h(\mathbf{x})]-\gamma\leq0,\\
\text{normal}, & \text{otherwise.}%
\end{cases}
\end{equation}

Introduce also the instance label $y_{s}$ which is $1$ if the training
instance $\mathbf{x}_{s}$ is anomalous, and $-1$ if it is normal. If labels
are not known, then prior values of labels can be determined by using the
original iForest.

We propose the following loss function:%
\begin{equation}
L\left(  h(\mathbf{x}_{s}),\mathbf{x}_{s},\mathbf{w}\right)  =\max\left(
0,y_{s}(E[h(\mathbf{x}_{s})]-\gamma)\right)  . \label{IForest_30}%
\end{equation}

It can be seen from (\ref{IForest_30}) that the loss function is $0$ if
$E[h(\mathbf{x}_{s})]-\gamma$ and $y_{s}$ have different signs, i.e., if the
decision about an anomalous (normal) instance coincides with the corresponding
label. Substituting (\ref{IForest_25}) into (\ref{IForest_30}), we rewrite
optimization problem (\ref{RF_Att_49}) as:
\begin{equation}
\mathbf{w}_{opt}=\arg\min_{\mathbf{w\in}\mathcal{W}}~\left[  \sum_{s=1}%
^{n}\max\left(  0,y_{s}\left(  \sum_{k=1}^{T}\alpha\left(  \mathbf{x}%
,\mathbf{A}_{k}(\mathbf{x)},\mathbf{w}\right)  \cdot h_{k}(\mathbf{x}%
)-\gamma\right)  \right)  \right]  . \label{IForest_32}%
\end{equation}

An important question is how to simplify the above problem to get a unique
solution and how to define the attention weights $\alpha\left(  \mathbf{x}%
,\mathbf{A}_{k}(\mathbf{x)},\mathbf{w}\right)  $ depending on the trainable
parameters $\mathbf{w}$. It can be done by using the Huber's $\epsilon
$-contamination model.

\subsection{Huber's contamination model}

We propose to use a simple representation of attention weights presented in
\cite{Utkin-Konstantinov-22}, which is based on applying the Huber's
$\epsilon$-contamination model \cite{Huber81}. The model is represented as a
set of discrete probability distributions $F$ of the form:
\begin{equation}
F=(1-\epsilon)\cdot P+\epsilon\cdot R,
\end{equation}
where $P=(p_{1},...,p_{T})$ is a discrete probability distribution
contaminated by another probability distribution denoted $R=(r_{1},...,r_{T})$
under condition that the probability distribution $R$ can be arbitrary; the
contamination parameter $\epsilon\in\lbrack0,1]$ controls the degree of the contamination.

The contaminating distribution $R$ is a point in the unit simplex with $T$
vertices denoted as $S(1,T)$. The distribution $F$ is a point in a small
simplex which belongs to the unit simplex. The size of the small simplex
depends on the hyperparameter $\epsilon$. If $\epsilon=1$, then the small
simplex coincides with the unit simplex. If $\epsilon=0$, then the the small
simplex is reduced to a single distribution $P$.

We propose to consider every element of $P$ as a result of the softmax
operation
\begin{equation}
p_{k}=\sigma\left(  -\frac{\left\Vert \mathbf{x}-\mathbf{A}_{k}(\mathbf{x)}%
\right\Vert ^{2}}{\omega}\right)  ,
\end{equation}
that is
\[
P=\left(  \sigma\left(  -\frac{\left\Vert \mathbf{x}-\mathbf{A}_{1}%
(\mathbf{x)}\right\Vert ^{2}}{\omega}\right)  ,...,\sigma\left(
-\frac{\left\Vert \mathbf{x}-\mathbf{A}_{T}(\mathbf{x)}\right\Vert ^{2}%
}{\omega}\right)  \right)  .
\]

Moreover, we propose to consider the distribution $R$ as the vector of
trainable parameters $\mathbf{w}$, that is
\[
R=\mathbf{w}=(w_{1},...,w_{T}).
\]

Hence, the attention weight $\alpha\left(  \mathbf{x},\mathbf{A}%
_{k}(\mathbf{x)},\mathbf{w}\right)  $ can be represented for every $k=1,...,T$
as follows:
\begin{equation}
\alpha\left(  \mathbf{x},\mathbf{A}_{k}(\mathbf{x)},\mathbf{w}\right)
=(1-\epsilon)\cdot\text{$\sigma$}\left(  -\frac{\left\Vert \mathbf{x}%
-\mathbf{A}_{k}(\mathbf{x)}\right\Vert ^{2}}{\omega}\right)  +\epsilon\cdot
w_{k}. \label{RSF_Att_90}%
\end{equation}

An important property of the above representation is that the attention weight
linearly depends on the trainable parameters, and the softmax operation
depends only on the hyperparameter $\omega$. The trainable parameters
$\mathbf{w=}(w_{1},...,w_{T})$ are restricted by the unit simplex $S(1,T)$
and, therefore, $\mathcal{W}=S(1,T)$. This implies that the constraints for
$\mathbf{w}$ are linear ($w_{i}\geq0$ and $w_{1}+...+w_{T}=1$).

\subsection{Loss function with the contamination model}

Let us substitute the obtained expression (\ref{RSF_Att_90}) for the attention
weight $\alpha\left(  \mathbf{x},\mathbf{A}_{k}(\mathbf{x)},\mathbf{w}\right)
$ into the objective function (\ref{IForest_32}). We get after simplification
\begin{equation}
\min_{\mathbf{w}\in S(1,T)}\sum_{s=1}^{n}\max\left(  0,D_{s}(\epsilon
,\omega)+y_{s}\epsilon\sum_{k=1}^{T}h_{k}(\mathbf{x}_{s})w_{k}\right)
\label{IForest_40}%
\end{equation}
where
\begin{equation}
D_{s}(\epsilon,\omega)=y_{s}(1-\epsilon)\sum_{k=1}^{T}\text{$\sigma$}\left(
-\frac{\left\Vert \mathbf{x}_{s}-\mathbf{A}_{k}(\mathbf{x}_{s}\mathbf{)}%
\right\Vert ^{2}}{\omega}\right)  -\gamma T.
\end{equation}

Let us introduce new variables
\begin{equation}
v_{s}=\max\left(  0,D_{s}(\epsilon,\omega)+y_{s}\epsilon\sum_{k=1}^{T}%
h_{k}(\mathbf{x}_{s})w_{k}\right)  .
\end{equation}

Then problem (\ref{IForest_40}) can be rewritten as follows:
\begin{equation}
\min\sum_{s=1}^{n}v_{s},
\end{equation}
subject to%
\begin{equation}
v_{s}\geq D_{s}(\epsilon,\omega)+y_{s}\epsilon\sum_{k=1}^{T}h_{k}%
(\mathbf{x}_{s})w_{k}, \label{IForest_47}%
\end{equation}%
\begin{equation}
v_{s}\geq0,\ s=1,...,n, \label{IForest_48}%
\end{equation}%
\begin{equation}
w_{1}+...+w_{T}=1,\ w_{k}\geq0,\ k=1,...,T. \label{IForest_49}%
\end{equation}

This is a linear optimization problem with the optimization variables
$w_{1},...,w_{T}$ and $v_{1},...,v_{n}$.

The optimization problem can be improved by adding a regularization term
$\left\Vert \mathbf{w}\right\Vert ^{2}$ with the hyperparameter $\lambda$
which controls the strength of the regularization. In this case, the
optimization problem becomes
\begin{equation}
\min\sum_{s=1}^{n}v_{s}+\lambda\left\Vert \mathbf{w}\right\Vert ^{2},
\label{IForest_50}%
\end{equation}
subject to (\ref{IForest_47}), (\ref{IForest_48}), (\ref{IForest_49}).

We get the standard quadratic programming problem whose solution does not meet
any difficulties.

\section{Numerical experiments}

The proposed attention-based iForest is studied by using synthetic and real
data and is compared with the original iForest. A brief introduction about
these datasets is given in Table \ref{t:anomaly_datasets} where $d$ is the
number of features, $n_{norm}$ and $n_{anom}$ are numbers of normal and
anomalous instances, respectively. 

Different values for hyper-parameters, including threshold $\tau$, the number
of trees in the forest, the contamination parameter $\epsilon$, the kernel
parameter $\omega$ have been tested, choosing those leading to the best
results. In particular, hyperparameter $\epsilon$ in ABIForest takes values
$0$, $0.25$, $0.5$, $0.75$, $1$; hyperparameter $\gamma$ changes from $0.5$ to
$0.7$; hyperparameter $\omega$ takes values $0.1$, $10$, $20$, $30$,
$40$\textbf{. }F1-score is used as a measure of the anomaly detection
accuracy. To evaluate the F1-score, a cross-validation with $100$ repetitions
is performed, where in each run, 66.7\% of data for training ($2n/3$) and
33.3\% for testing ($n/3$) are randomly selected. Numerical results are
presented in tables where the best results are shown in bold. %

\begin{table}[tbp] \centering
\caption{A brief introduction about datasets}%
\begin{tabular}
[c]{cccc}\hline
Dataset & $n_{norm}$ & $n_{anom}$ & $d$\\\hline
Circle (synthetic) & $1000$ & $200$ & $2$\\\hline
Normal dataset (synthetic) & $1000$ & $50$ & $2$\\\hline
Credit & $1500$ & $400$ & $30$\\\hline
Ionosphere & $225$ & $126$ & $33$\\\hline
Arrhythmia & $386$ & $66$ & $18$\\\hline
Mulcross & $1800$ & $400$ & $4$\\\hline
Http & $500$ & $50$ & $3$\\\hline
Pima & $500$ & $268$ & $8$\\\hline
\end{tabular}
\label{t:anomaly_datasets}%
\end{table}%

\subsection{Synthetic datasets}

\begin{figure}
[ptb]
\begin{center}
\includegraphics[
height=2.065in,
width=4.6295in
]%
{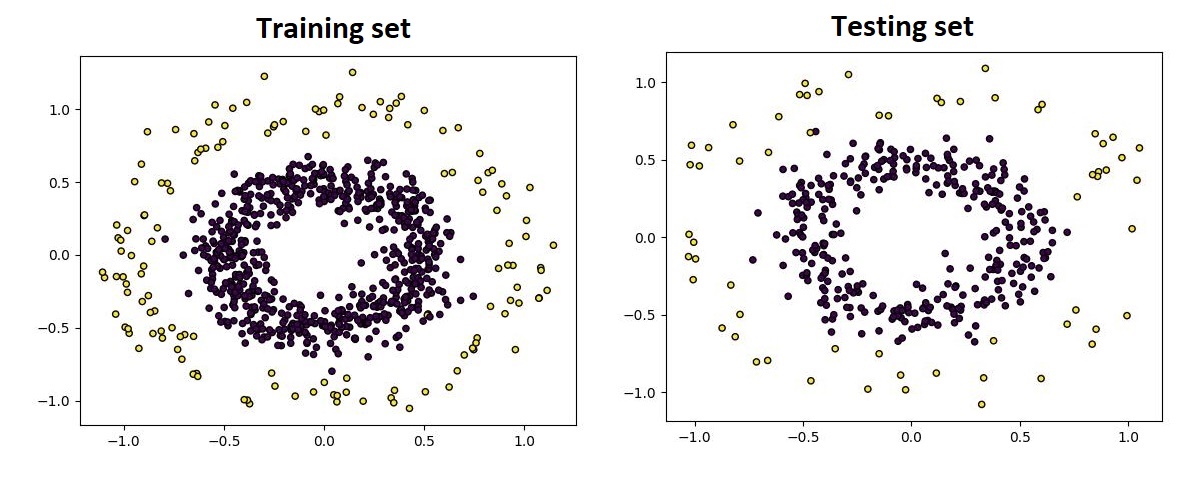}%
\caption{Points from the Circle dataset}%
\label{f:Circle_1}%
\end{center}
\end{figure}

The first synthetic dataset used for numerical experiments is the
\textit{Circle} dataset. Its points are divided into two parts concentrated
around small and large circles as it is shown in Fig. \ref{f:Circle_1} where
the training and testing sets are depicted in the left and rights pictures,
respectively. In order to optimize the model parameters in numerical
experiments, we perform a cross-validation. The Gaussian noise with the
standard deviation $0.1$ is added to data for all experiments.%

\begin{figure}
[ptb]
\begin{center}
\includegraphics[
height=2.0166in,
width=4.7729in
]%
{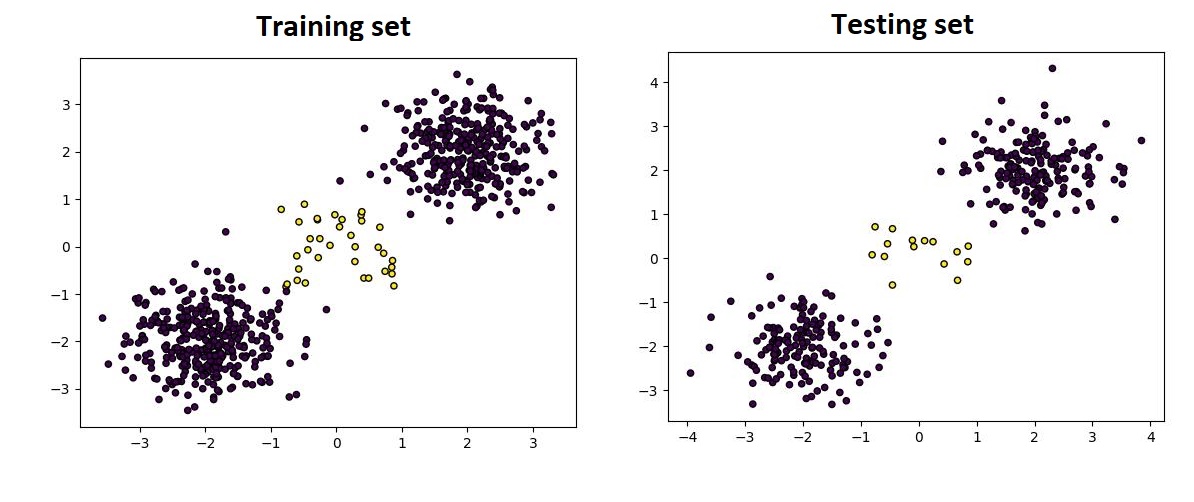}%
\caption{Points from the Normal dataset}%
\label{f:Normal_1}%
\end{center}
\end{figure}

The second synthetic dataset (the \textit{Normal} dataset) contains points
generated from the normal distributions with two expectations $(-2,-2)$ and
$(2,2)$. Anomalies are generated from the uniform distribution in interval
$[-1,1]$. Training and testing sets are depicted in Fig. \ref{f:Normal_1}.%

First, we study the Circle dataset. F1-score measures obtained for ABIForest
are shown in Table \ref{t:IForest_2} where the F1-score is presented as the
function of hyperparameters $\epsilon$ and $\tau$ by the number of trees in
the isolation forest $T=150$. It is interesting to note that ABIForest is
sensitive to changes of $\tau$ whereas $\epsilon$ does not significantly
impact on results. For comparison purposes, F1-score measures of the original
iForest as a function of the number $T$ of trees and hyperparameter $\tau$ are
shown in Table \ref{t:IForest_1}. It can be seen from Table \ref{t:IForest_1}
that the largest value of the F1-score is achieved by $150$ trees in the
forest and by $\tau=0.5$. One can also see from Tables \ref{t:IForest_2} and
\ref{t:IForest_1} that ABIForest provides results which outperform the same
results of the original iForest.%

\begin{table}[tbp] \centering
\caption{F1-score of ABIForest consisting of $T=150$ trees as a function of hyperparameters $\tau $ and $\epsilon $  for the Circle dataset by $\omega =20$}%
\begin{tabular}
[c]{lccc}\hline
$\epsilon$ & \multicolumn{3}{c}{$\tau$}\\\hline
& $0.5$ & $0.6$ & $0.7$\\\hline
$0.0$ & $0.276$ & $0.973$ & $0.236$\\\hline
$0.25$ & $0.2749$ & $0.975$ & $0.162$\\\hline
$0.5$ & $0.273$ & $\mathbf{0.978}$ & $0.100$\\\hline
$0.75$ & $0.273$ & $0.975$ & $0.062$\\\hline
$1.0$ & $0.271$ & $0.973$ & $0.037$\\\hline
\end{tabular}
\label{t:IForest_2}%
\end{table}%
%

\begin{table}[tbp] \centering
\caption{F1--score of the original iForest as a function of the number $T$ of trees and hyperparameter $\tau $ for the Circle dataset}%
\begin{tabular}
[c]{lccccc}\hline
$\tau$ & \multicolumn{5}{c}{$T$}\\\hline
& $5$ & $15$ & $25$ & $50$ & $150$\\\hline
$0.3$ & $0.270$ & $0.270$ & $0.270$ & $0.270$ & $0.270$\\\hline
$0.4$ & $0.286$ & $0.273$ & $0.271$ & $0.270$ & $0.270$\\\hline
$0.5$ & $0.729$ & $0.864$ & $0.899$ & $0.906$ & $\mathbf{0.920}$\\\hline
$0.6$ & $0.639$ & $0.603$ & $0.598$ & $0.603$ & $0.606$\\\hline
\end{tabular}
\label{t:IForest_1}%
\end{table}%

Similar numerical experiments with the Normal dataset are presented in Tables
\ref{t:IForest_4} and \ref{t:IForest_3}. We can again see that ABIForest
outperforms the iForest, namely the best value of the F1-score provided by the
iForest is $0.252$ whereas the best value of the F1--score for ABIForest is
$0.413$, and this result is obtained by $\omega=20$.%

\begin{table}[tbp] \centering
\caption{F1-score of ABIForest consisting of $T=150$ trees as a function of hyperparameters $\tau $ and $\epsilon $  for the Normal dataset by $\omega =20$}%
\begin{tabular}
[c]{lccc}\hline
$\epsilon$ & \multicolumn{3}{c}{$\tau$}\\\hline
& $0.5$ & $0.6$ & $0.7$\\\hline
$0.0$ & $0.099$ & $0.410$ & $0.0$\\\hline
$0.25$ & $0.147$ & $0.410$ & $0.162$\\\hline
$0.5$ & $0.177$ & $\mathbf{0.413}$ & $0.0$\\\hline
$0.75$ & $0.176$ & $0.412$ & $0.0$\\\hline
$1.0$ & $0.178$ & $0.408$ & $0.0$\\\hline
\end{tabular}
\label{t:IForest_4}%
\end{table}%
%

\begin{table}[tbp] \centering
\caption{F1--score of the original iForest as a function of the number $T$ of trees and hyperparameter $\tau $ for the Normal dataset}%
\begin{tabular}
[c]{lccccc}\hline
$\tau$ & \multicolumn{5}{c}{$T$}\\\hline
& $5$ & $15$ & $25$ & $50$ & $150$\\\hline
$0.3$ & $0.082$ & $0.082$ & $0.082$ & $0.082$ & $0.082$\\\hline
$0.4$ & $0.088$ & $0.083$ & $0.083$ & $0.082$ & $0.082$\\\hline
$0.5$ & $0.220$ & $0.248$ & $0.249$ & $0.250$ & $\mathbf{0.252}$\\\hline
$0.6$ & $0.191$ & $0.141$ & $0.091$ & $0.040$ & $0.021$\\\hline
\end{tabular}
\label{t:IForest_3}%
\end{table}%

\begin{figure}
[ptb]
\begin{center}
\includegraphics[
height=2.6516in,
width=3.4932in
]%
{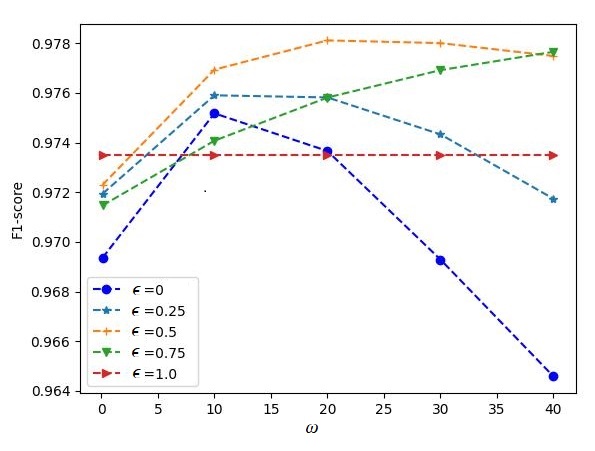}%
\caption{F1--score measures as functions of the softmax hyperparameter
$\omega$ for diffrent contamination parameters $\epsilon$ for the Circle
dataset}%
\label{fig:circle_tau_eps}%
\end{center}
\end{figure}

\begin{figure}
[ptb]
\begin{center}
\includegraphics[
height=2.6745in,
width=3.5284in
]%
{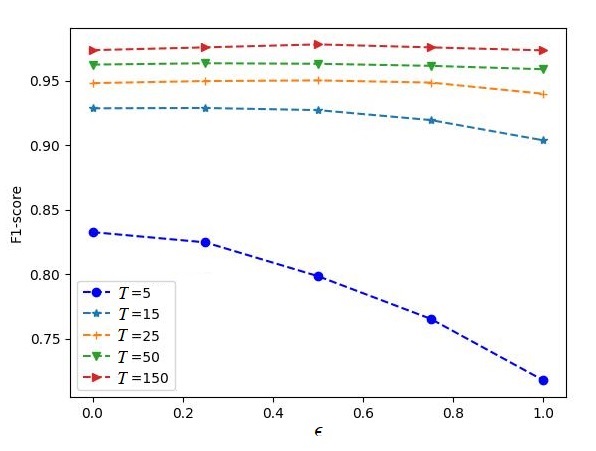}%
\caption{F1-score measures as functions of the contamination parameter
$\omega$ for different numbers of trees in iForest $T$ for the Circle dataset}%
\label{fig:circle_eps_T}%
\end{center}
\end{figure}

Fig. \ref{fig:circle_tau_eps} illustrates how the F1-score depends on
hyperparameter $\tau$ for the Circle dataset. The corresponding functions are
depicted for different contamination parameters $\epsilon$ and obtained for
the case of $T=150$ trees in the iForest. It can be seen from Fig.
\ref{fig:circle_tau_eps} that the largest value of the F1-score is achieved by
$\omega=20$ and $\epsilon=0.5$. It can also be seen from the results in Fig.
\ref{fig:circle_tau_eps} that the F1-score significantly depends on
hyperparameter $\omega$ especially for small values of $\epsilon$. F1-score
measures as functions of the contamination parameter $\omega$ for different
numbers of trees in the iForest $T$ for the Circle dataset obtained by
hyperparameters $\gamma=0.6$ and $\omega=20$ are depicted in Fig.
\ref{fig:circle_eps_T}.%

%

\begin{figure}
[ptb]
\begin{center}
\includegraphics[
height=1.58in,
width=5.6126in
]%
{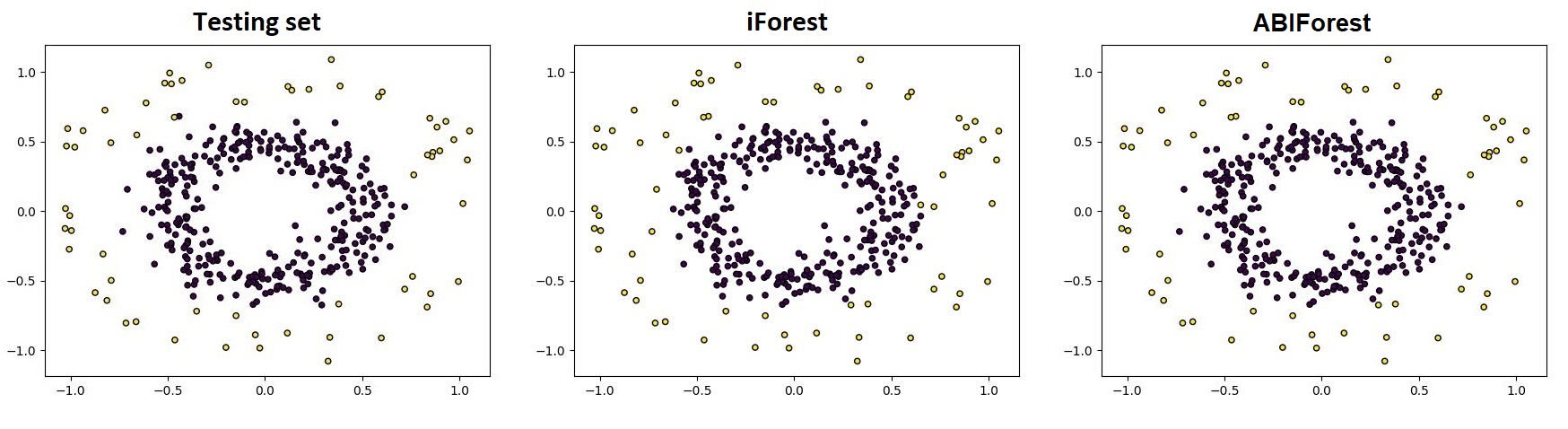}%
\caption{Comparison of the generated testing set for the Circle dataset (the
left picture), predictions obtained by iForest (the central picture),
predictions obtained by ABIForest (the right picture)}%
\label{f:circle_res_test_1000}%
\end{center}
\end{figure}

Fig. \ref{f:circle_res_test_1000} illustrates comparison results between the
iForest and ABIForest on the basis of the testing set which is depicted in the
left picture of Fig. \ref{f:circle_res_test_1000}. Predictions obtained by the
iForest consisting of $150$ trees by $\tau=0.5$ are depicted in the central
picture. Predictions obtained by ABIForest by $\epsilon=0.5$, $\tau=0.6$ and
$\omega=0.1$ are shown in the right picture. One can see from Fig.
\ref{f:circle_res_test_1000} that some points in the central picture are
incorrectly identified as anomalous ones whereas ABIForest is correctly
classified them as normal instance. Fig. \ref{f:circle_res_test_1000} should
not be considered as the single realization which defines the F1-score. It is
one of many cases corresponding to different generations of testing sets,
therefore, the numbers of normal and anomalous instances can be different in
each realization.%

\begin{figure}
[ptb]
\begin{center}
\includegraphics[
height=2.8108in,
width=3.4853in
]%
{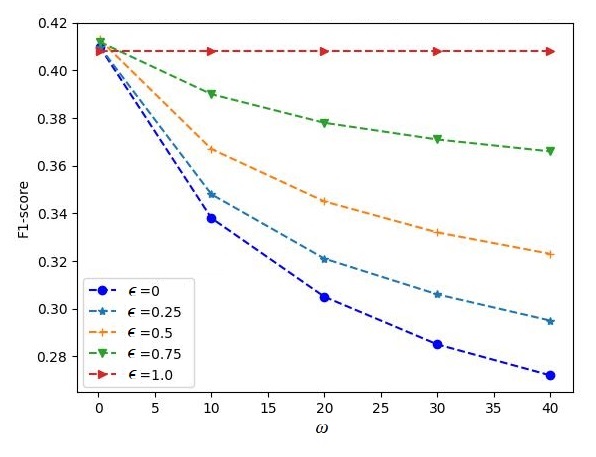}%
\caption{F1-score measures as functions of the softmax hyperparameter $\omega$
for different contamination parameters $\epsilon$ and for the Normal dataset}%
\label{fig:normal_tau_eps}%
\end{center}
\end{figure}

\begin{figure}
[ptb]
\begin{center}
\includegraphics[
height=2.6727in,
width=3.5223in
]%
{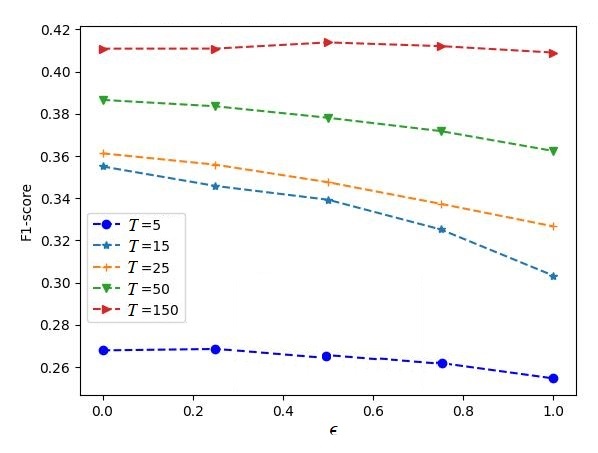}%
\caption{F1-score measures as functions of the contamination parameter
$\omega$ for different numbers of trees in iForest $T$ for the Circle dataset}%
\label{fig:normal_eps_T}%
\end{center}
\end{figure}

Similar dependencies for the Normal dataset are shown in Figs.
\ref{fig:normal_tau_eps} and \ref{fig:normal_eps_T}. However, it follows from
Fig. \ref{fig:normal_tau_eps} that the largest values of the F1-score are
achieved for $\epsilon=0$. This implies that the main contribution into the
attention weights is caused by the softmax operation. F1-score measures shown
in Fig. \ref{fig:normal_eps_T} are obtained by hyperparameters $\gamma=0.6$
and $\omega=20$.%

%

\begin{figure}
[ptb]
\begin{center}
\includegraphics[
height=1.6903in,
width=6.0639in
]%
{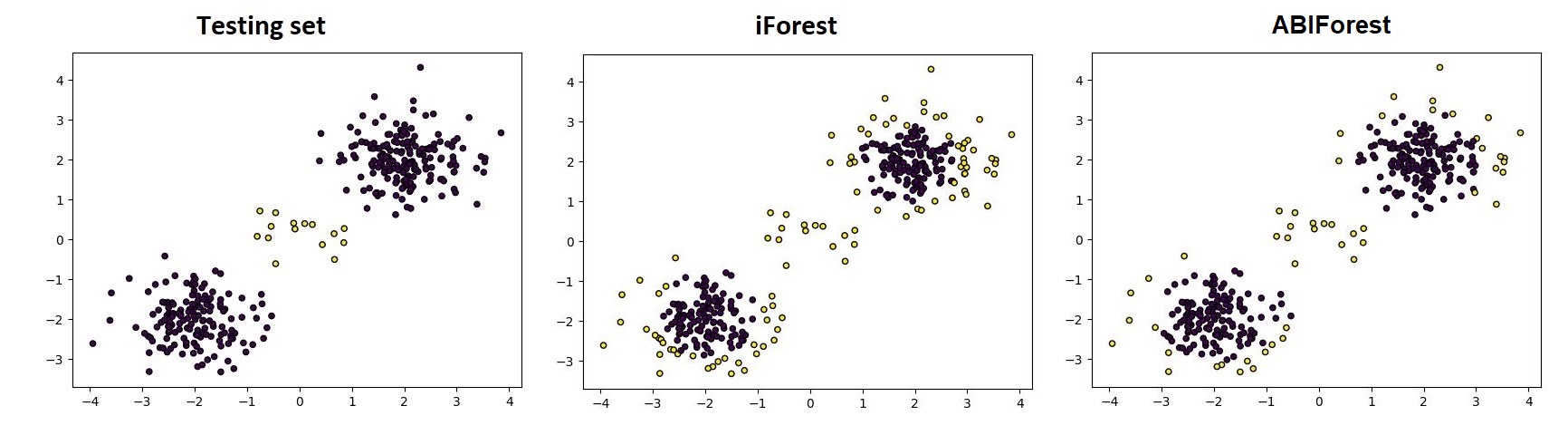}%
\caption{Comparison of the generated testing set for the Normal dataset (the
left picture), predictions obtained by iForest (the central picture),
predictions obtained by ABIForest (the right picture)}%
\label{f:normal_test_if_ABIForest}%
\end{center}
\end{figure}

Comparison results between the iForest and ABIForest for the Normal dataset
are shown in Fig. \ref{f:normal_test_if_ABIForest} where a realization of the
testing set, predictions of the iForest and ABIForest are shown in the left,
central, right pictures, respectively. Predictions are obtained by means of
the iForest consisting of $150$ trees by $\tau=0.5$ and by ABIForest
consisting of the same number of trees by $\epsilon=0.5$, $\tau=0.6$ and
$\omega=0.1$.%

\begin{figure}
[ptb]
\begin{center}
\includegraphics[
height=1.9691in,
width=4.5697in
]%
{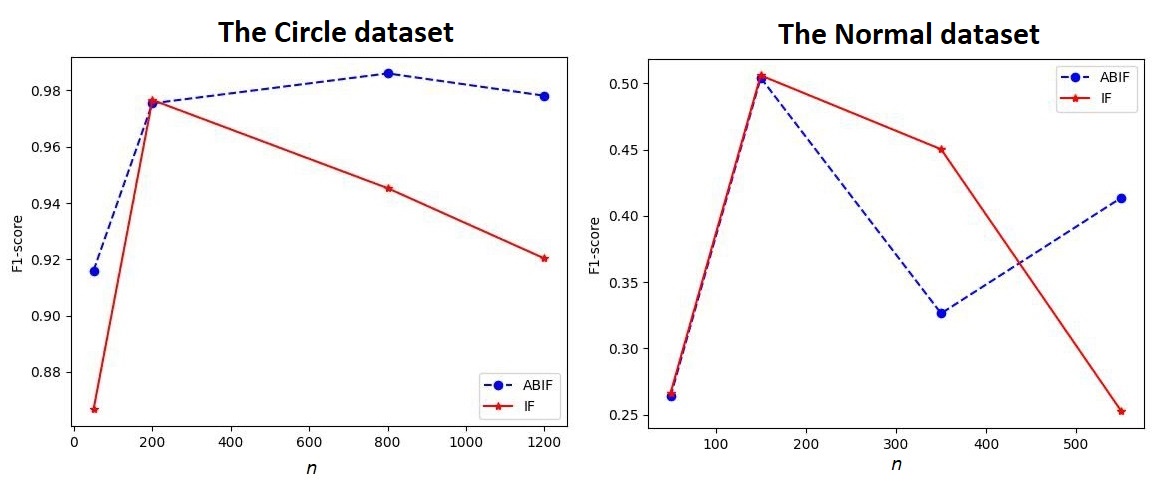}%
\caption{Illustration how the F1-score measures of iForest and ABIForest
depend on numbers of training data for the Circle dataset (the left picture)
and the Normal dataset (the right picture)}%
\label{f:circle_normal_n}%
\end{center}
\end{figure}

Another interesting question is how the prediction accuracy of ABIForest
depends on the size of training data. The corresponding results for synthetic
datasets are shown in Fig. \ref{f:circle_normal_n} where the solid and dashed
lines correspond to the F1-score of iForest and ABIForest, respectively.
Numbers of trees in all experiments are taken $T=150$. The same results in the
numerical form are also given in Table \ref{t:IForest_5}. It can be seen from
Fig. \ref{f:circle_normal_n} for the Circle dataset that the F1-score of the
iForest decreases with increase of the number of training data after $n=200$.
This is because the number of trees ($T=150$) is fixed and trees cannot be
improved. This effect has been discussed in \cite{Liu-Ting-Zhou-2012} where
problems of swamping and masking were studied. Authors
\cite{Liu-Ting-Zhou-2012} considered the subsampling to overcome these
problems. One can see from Fig. \ref{f:circle_normal_n} that ABIForest copes
with this difficulty. Another behavior of ABIForest can be observed for the
Normal dataset which is characterized by two clusters of normal points. In
this case, the F1-score decreases as $n$ increases and then increases with $n$.%

%

\begin{table}[tbp] \centering
\caption{F1-score measures of the original iForest and ABIForest as functions of the training data number $n$ for the Circle and Normal datasets}%
\begin{tabular}
[c]{lcccc}\hline
\multicolumn{5}{l}{The Circle dataset}\\\hline
$n$ & $50$ & $200$ & $800$ & $1200$\\\hline
iForest & $0.867$ & $\mathbf{0.977}$ & $0.945$ & $0.920$\\\hline
ABIForest & $0.916$ & $0.975$ & $\mathbf{0.986}$ & $0.978$\\\hline
\multicolumn{5}{l}{The Normal dataset}\\\hline
$n$ & $50$ & $150$ & $350$ & $550$\\\hline
iForest & $0.266$ & $\mathbf{0.506}$ & $0.450$ & $0.253$\\\hline
ABIForest & $0.264$ & $\mathbf{0.512}$ & $0.326$ & $0.413$\\\hline
\end{tabular}
\label{t:IForest_5}%
\end{table}%

\subsection{Real dataset}

The first real dataset, used in numerical experiments and called the
\textit{Credit}
dataset\footnote{https://www.kaggle.com/datasets/mlg-ulb/creditcardfraud}.
According to the dataset description, it contains transactions made by credit
cards in September 2013 by European cardholders with $492$ frauds out of
$284807$ transactions. We use only $1500$ normal instances and $400$ anomalous
ones, which are randomly selected from the whole Credit dataset. The second
dataset, called the \textit{Ionosphere}
dataset\footnote{https://www.kaggle.com/datasets/prashant111/ionosphere}, is a
collection of radar returns from the ionosphere. The next dataset is called
the \textit{Arrhythmia}
dataset\footnote{https://www.kaggle.com/code/medahmedkrichen/arrhythmia-classification}%
. The smallest classes with numbers 3, 4, 5, 7, 8, 9, 14, 15 are combined to
form outliers in the Arrhythmia dataset. The \textit{Mulcross}
dataset\footnote{https://github.com/dple/Datasets} is generated from a
multi-variate normal distribution with two dense anomaly clusters. We use
$1800$ normal and $400$ anomalous instances. The \textit{Http}
dataset\footnote{http://odds.cs.stonybrook.edu/http-kddcup99-dataset/} is used
in \cite{Liu-Ting-Zhou-2012} for studying iForest. The \textit{Pima}
dataset\footnote{https://github.com/dple/Datasets} aims to predict whether or
not a patient has diabetes. Datasets Credit, Mulcross, Http are reduced to
simplify experiments.

Numerical results are shown in Table \ref{t:IForest_7}. ABIForest is presented
in Table \ref{t:IForest_7} by hyperparameters $\epsilon$, {$\tau$, }${\omega}$
and the F1-score. iForest is presented by hyperparameter {$\tau$ and }the
corresponding F1-score. Hyperparameters leading to the largest F1-score are
presented in Table \ref{t:IForest_7}. It can be seen from Table
\ref{t:IForest_7} that ABIForest provides outperforming results for five from
six datasets. It is also interesting to point out that optimal values of
hyperparameter $\epsilon$ for two datasets Ionosphere and Mullcross are equal
to $0$. This implies that attention weights are entirely determined by the
softmax operation (see (\ref{RSF_Att_90})). A contrary case is when
{$\epsilon_{opt}=1$. In this case, the softmax operations as well as their
parameter }${\omega}${ are not used, and the attention weights are }entirely
determined by parameters $\mathbf{w}$ which can be regarded as weights of trees.%

\begin{table}[tbp] \centering
\caption{F1-score measures of ABIForest consisting of $T=150$ trees for different real datasets by optimal values of $\tau $, $\epsilon $, $\omega $ and F1-score measures of iForest by optimal values of $\tau $}%
\begin{tabular}
[c]{ccccccc}\hline
& \multicolumn{4}{c}{ABIForest} & \multicolumn{2}{c}{iForest}\\\hline
Dataset & {$\epsilon_{opt}$} & {$\tau_{opt}$} & ${\omega}_{opt}$ & {F1} &
{$\tau_{opt}$} & {F1}\\\hline
Credit & $0.25$ & $0.55$ & $0.1$ & $\mathbf{0.911}$ & $0.4$ & $0.836$\\\hline
Ionosphere & $0.0$ & $0.4$ & $0.1$ & $\mathbf{0.693}$ & $0.45$ &
$0.684$\\\hline
Arrhythmia & $1.0$ & $0.45$ & $-$ & $\mathbf{0.481}$ & $0.4$ & $0.479$\\\hline
Mullcross & $0.0$ & $0.6$ & $0.1$ & $0.507$ & $0.5$ & $\mathbf{0.516}$\\\hline
Http & $0.75$ & $0.55$ & $0.1$ & $\mathbf{0.843}$ & $0.5$ & $0.720$\\\hline
Pima & $0.75$ & $0.45$ & $30$ & $\mathbf{0.553}$ & $0.4$ & $0.540$\\\hline
\end{tabular}
\label{t:IForest_7}%
\end{table}%

\begin{figure}
[ptb]
\begin{center}
\includegraphics[
height=2.3921in,
width=5.4959in
]%
{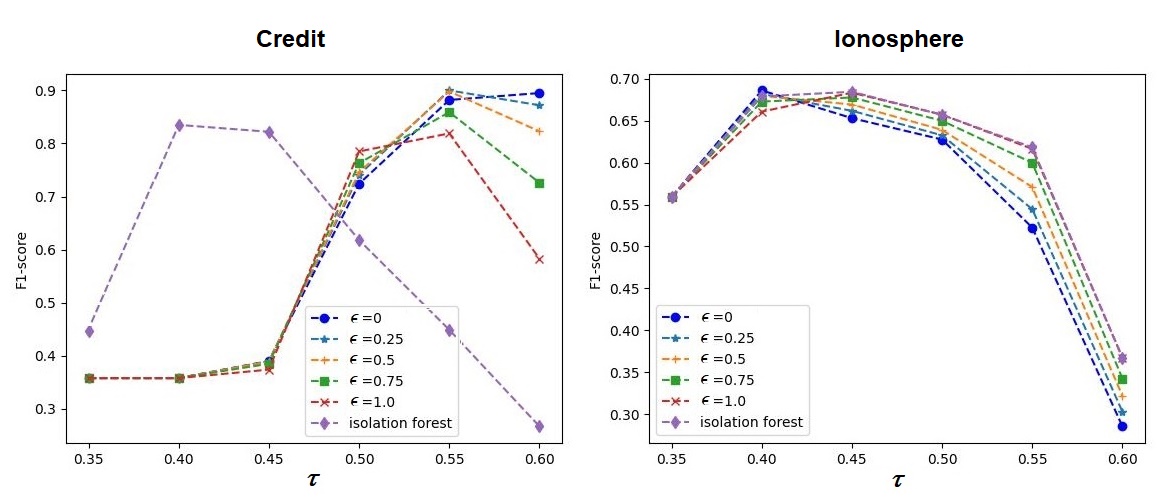}%
\caption{Comparison of iForest and ABIForest by different thresholds $\tau$
and by different contamination parameter $\epsilon$ for the Credit (the left
picture) and Ionosphere (the right picture) datasets}%
\label{f:credit_ion_tau}%
\end{center}
\end{figure}

\begin{figure}
[ptb]
\begin{center}
\includegraphics[
height=2.3583in,
width=5.5158in
]%
{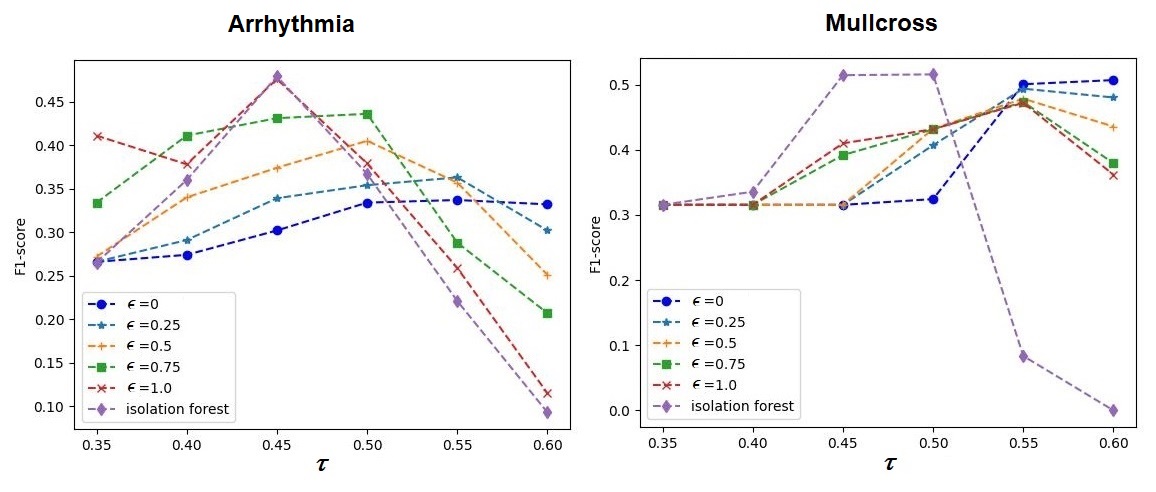}%
\caption{Comparison of iForest and ABIForest by different thresholds $\tau$
and by different contamination parameter $\epsilon$ for the Arrhithmia (the
left picture) and Mullcross (the right picture) datasets}%
\label{f:arrythmia_mull_tau}%
\end{center}
\end{figure}

\begin{figure}
[ptb]
\begin{center}
\includegraphics[
height=2.3237in,
width=5.5045in
]%
{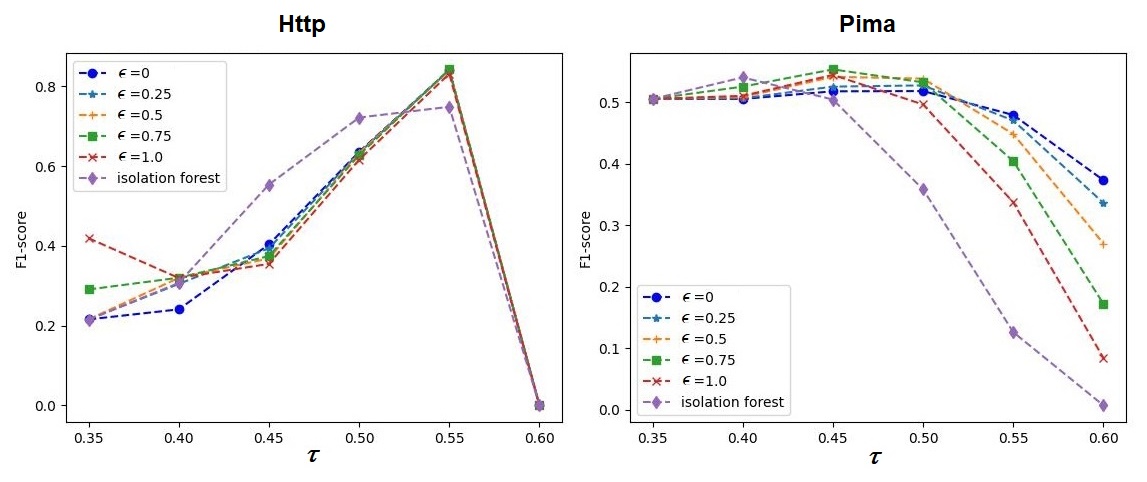}%
\caption{Comparison of iForest and ABIForest by different thresholds $\tau$
and by different contamination parameter $\epsilon$ for the Http (the left
picture) and Pima (the right picture) datasets}%
\label{f:http_pima_tau}%
\end{center}
\end{figure}

It is interesting to study how hyperparameter $\tau$ impacts on the
performance of ABIForest and iForest. The corresponding dependencies are
depicted in Figs. \ref{f:credit_ion_tau}-\ref{f:http_pima_tau}. The comparison
results are obtained under condition of optimal values of {$\epsilon$ and
}${\omega}$ given in Table \ref{t:IForest_7}. One can see from Fig.
\ref{f:credit_ion_tau} that {$\tau$ differently impacts on performances of
}ABIForest and iForest for the Credit dataset whereas the corresponding
dependencies scarcely differ for the Ionosphere dataset. This peculiarity is
caused by the optimal values of the contamination parameter {$\epsilon$. It
can be seen from }Table \ref{t:IForest_7} that {$\tau_{opt}=0$ for the
}Ionosphere dataset. This implies that the attention weights are determined
only by the softmax operations which weakly impact on the model performance
and whose values are close to $1/T$. Moreover, the Ionosphere dataset is one
of the smallest datasets with a large number of anomalous instances (see Table
\ref{t:anomaly_datasets}). Therefore, additional learnable parameters may lead
to overfitting. This is a reason why the optimal hyperparameter $\epsilon$
does not impact on the model performance. It is also interesting to note that
the optimal value of the contamination parameter for the Mullcross dataset is
$0$ (see Table \ref{t:IForest_7}). However, one can see quite different
dependencies from the right picture of Fig. \ref{f:arrythmia_mull_tau}. This
is caused by a large impact of the softmax operations whose values are far
from $1/T$, and they provide results different from iForest.

Generally, one can see from Figs. \ref{f:credit_ion_tau}-\ref{f:http_pima_tau}
that models strongly depend on hyperparameters $\tau$ and $\epsilon$. Most
dependencies illustrate that there is an optimal value of $\tau$ for each
case, which is close to $0.5$ for iForest as well as for ABIForest. The same
can be said about contamination parameter $\epsilon$.%

%

%

\section{Concluding remark}

A new modification of iForest using the attention mechanism has been proposed.
Let us focus on advantages and disadvantages of the modification.

\textbf{Advantages:}

\begin{enumerate}
\item ABIForest is very simple from the computation point of view because, in
contrast to the attention-based neural network, the attention weights in
ABIForest are trained by solving the standard quadratic optimization problem.
The modification avoids gradient-based algorithms to compute optimal learnable
attention parameters.

\item ABIForest is a flexible model which can be simply modified. There are
several components of ABIForest, which can be changed to improve the model
performance. First, different kernels can be used instead of the Gaussian
kernel considered above. Second, there are statistical models \cite{Walley91}
different from the Huber's $\epsilon$-contamination model, which can also be
used in ABIForest. Third, the attention weights can be associated with some
subsets of trees, including intersecting subsets. In this case, the number of
trainable parameters can be reduced to avoid overfitting. Fourth, paths in
trees can be also attended, for example, by assigning attention weights to
each branch in every path. Fifth, the multi-head attention can be applied to
iForest in order to improve the model, for example, by changing hyperparameter
$\omega$ of the softmax. Sixth, the distance between the instance $\mathbf{x}$
and all instances, which fall in the same leaf as $\mathbf{x}$, can be defined
differently. The above improvements can be regarded as directions for further research.

\item The attention model is trained after the forest building. This implies
that we do not need to rebuild iForest to achieve a higher accuracy.
Hyperparameters are tuned without rebuilding iForest. Moreover, we can apply
various modifications and extensions of iForest and incorporate the attention
mechanism in the same way as it is carried out with the original iForest.

\item ABIForest allows us to get interpretation answering the question why an
instance is anomalous. This can be done by analyzing isolation trees with the
largest attention weights.

\item ABIForest is perfectly deals with tabular data.

\item It follows from numerical experiments that ABIForest improves the
iForest performance for many datasets.
\end{enumerate}

\textbf{Disadvantages:}

\begin{enumerate}
\item The main disadvantage is that ABIForest has additionally three
hyperparameters: contamination parameter $\epsilon$, hyperparameter of the
softmax operation $\omega$, regularization hyperparameter $\lambda$. We do not
include threshold $\tau$ which is also used in iForest. Additional
hyperparameters lead to significant increase of the validation time.

\item Some additional time is required to solve the optimization problem
(\ref{IForest_32}).

\item In contrast to iForest, ABIForest is a supervised model. It requires to
have labels of data (normal or anomalous) in order to determine a criteria of
the optimization, in particular, to determine the optimization problem
(\ref{IForest_32}).
\end{enumerate}

In spite of the disadvantages, ABIForest can be viewed as the first version
for incorporating the attention mechanism into iForest which has illustrated
outperforming results. The following modifications resolving the above
disadvantages are interesting directions for further research.

\bibliographystyle{unsrt}
\bibliography{Anomaly_det,Attention,Boosting,Classif_bib,Cluster_bib,Deep_Forest,Imprbib,MYBIB,MYUSE,Robots}

\end{document}